\documentclass[conference]{IEEEtran}
\IEEEoverridecommandlockouts
\usepackage{cite}
\usepackage{amsmath,amssymb,amsfonts}
\usepackage{algorithmic}
\usepackage{graphicx}
\usepackage{textcomp}
\usepackage{xcolor}
\usepackage{textcomp}
\usepackage{xcolor}
\usepackage{tabularx}  

\usepackage[hidelinks,colorlinks=true]{hyperref}
\hypersetup{hidelinks}
\usepackage{float}  
\usepackage{multirow}
\usepackage{tcolorbox}
\usepackage{array}
\def\BibTeX{{\rm B\kern-.05em{\sc i\kern-.025em b}\kern-.08em
    T\kern-.1667em\lower.7ex\hbox{E}\kern-.125emX}}
\begin{document}

\title{DBF-UNet: A Two-Stage Framework for Carotid Artery Segmentation with Pseudo-Label Generation}
\author{\IEEEauthorblockN{Haoxuan Li\textsuperscript{1,*}, Wei Song\textsuperscript{2,*},Aofan Liu\textsuperscript{3},Peiwu Qin\textsuperscript{1,†}}
    \IEEEauthorblockA{
    \textsuperscript{1} Shenzhen International Graduate School, Tsinghua University, Shenzhen, China\\
    \textsuperscript{2} School of Automation, Guangdong University of Technology, Guangzhou, China\\
    \textsuperscript{3} School of Electronic and Computer Engineering, Peking University, Shenzhen, China\\
    \textsuperscript{*}These authors contributed equally to this work. \\
    \textsuperscript{†}Corresponding author: pwqin@sz.tsinghua.edu.cn
    }
}
\maketitle
\begin{abstract}
Medical image analysis faces significant challenges due to limited annotation data, particularly in three-dimensional carotid artery segmentation tasks, where existing datasets exhibit spatially discontinuous slice annotations with only a small portion of expert-labeled slices in complete 3D volumetric data. To address this challenge, we propose a two-stage segmentation framework. First, we construct continuous vessel centerlines by interpolating between annotated slice centroids and propagate labels along these centerlines to generate interpolated annotations for unlabeled slices. The slices with expert annotations are used for fine-tuning SAM-Med2D, while the interpolated labels on unlabeled slices serve as prompts to guide segmentation during inference. In the second stage, we propose a novel Dense Bidirectional Feature Fusion UNet (DBF-UNet). This lightweight architecture achieves precise segmentation of complete 3D vascular structures. The network incorporates bidirectional feature fusion in the encoder and integrates multi-scale feature aggregation with dense connectivity for effective feature reuse. Experimental validation on public datasets demonstrates that our proposed method effectively addresses the sparse annotation challenge in carotid artery segmentation while achieving superior performance compared to existing approaches. The source code is available at \url{https://github.com/Haoxuanli-Thu/DBF-UNet}.
\end{abstract}
\section{Introduction}
Atherosclerosis, a chronic inflammatory disease that severely threatens global human health, is characterized by lipid plaque deposition, smooth muscle cell proliferation, and extracellular matrix accumulation in arterial walls, predominantly affecting the carotid arteries that supply blood to the brain~\cite{libby2009inflammation}. Magnetic Resonance (MR) black-blood vessel wall imaging (BB-VWI) has emerged as a powerful diagnostic tool, effectively visualizing both normal and pathological arterial vessel walls while providing critical evidence for atherosclerosis characterization. This technology has demonstrated exceptional capability in revealing vessel wall abnormalities and quantifying atherosclerotic burden, making it invaluable for clinical diagnosis and treatment planning ~\cite{hu2022label}.

Despite the clinical significance of carotid artery segmentation, current manual annotation approaches face critical challenges in clinical practice. The process demands extensive time investment from experienced radiologists and is highly susceptible to inter-observer variability. This variability stems from differences in clinical expertise and subjective interpretation of image features, potentially compromising the reliability of vessel analysis. Moreover, the complex geometric structures of atherosclerotic lesions further complicate the accurate delineation of vessel boundaries. These limitations highlight the urgent need for automated segmentation solutions to enhance diagnostic efficiency and accuracy~\cite{liu2021extraction}.

Deep learning-based image segmentation algorithms present novel solutions to these challenges~\cite{ronneberger2015u,hatamizadeh2022unetr,cao2022swin}. These algorithms can automatically learn and extract image features, significantly reducing manual intervention while improving the accuracy and precision of carotid artery segmentation. By learning reliable feature representations from large-scale medical imaging datasets, these algorithms efficiently complete segmentation tasks, reducing diagnostic time, alleviating workload for physicians, and potentially lowering healthcare costs. Consequently, deep learning-based carotid artery segmentation has emerged as a critical research focus and technical challenge in medical image processing, holding both theoretical value and practical significance, with anticipated crucial roles in future medical diagnosis and treatment.

Automated carotid artery segmentation currently faces two challenges: first, high-quality annotation data often lacks spatial continuity, limiting the effectiveness of 3D segmentation model training; second, interpolated annotations become unreliable at vessel bifurcations and pathological regions where carotid arteries undergo significant structural changes, making accurate vessel delineation difficult in these critical areas.

Addressing these challenges through the development of robust automated segmentation algorithms holds significant clinical value and research importance.
\begin{figure}[htbp]
\centering
\includegraphics[width=0.47\textwidth]{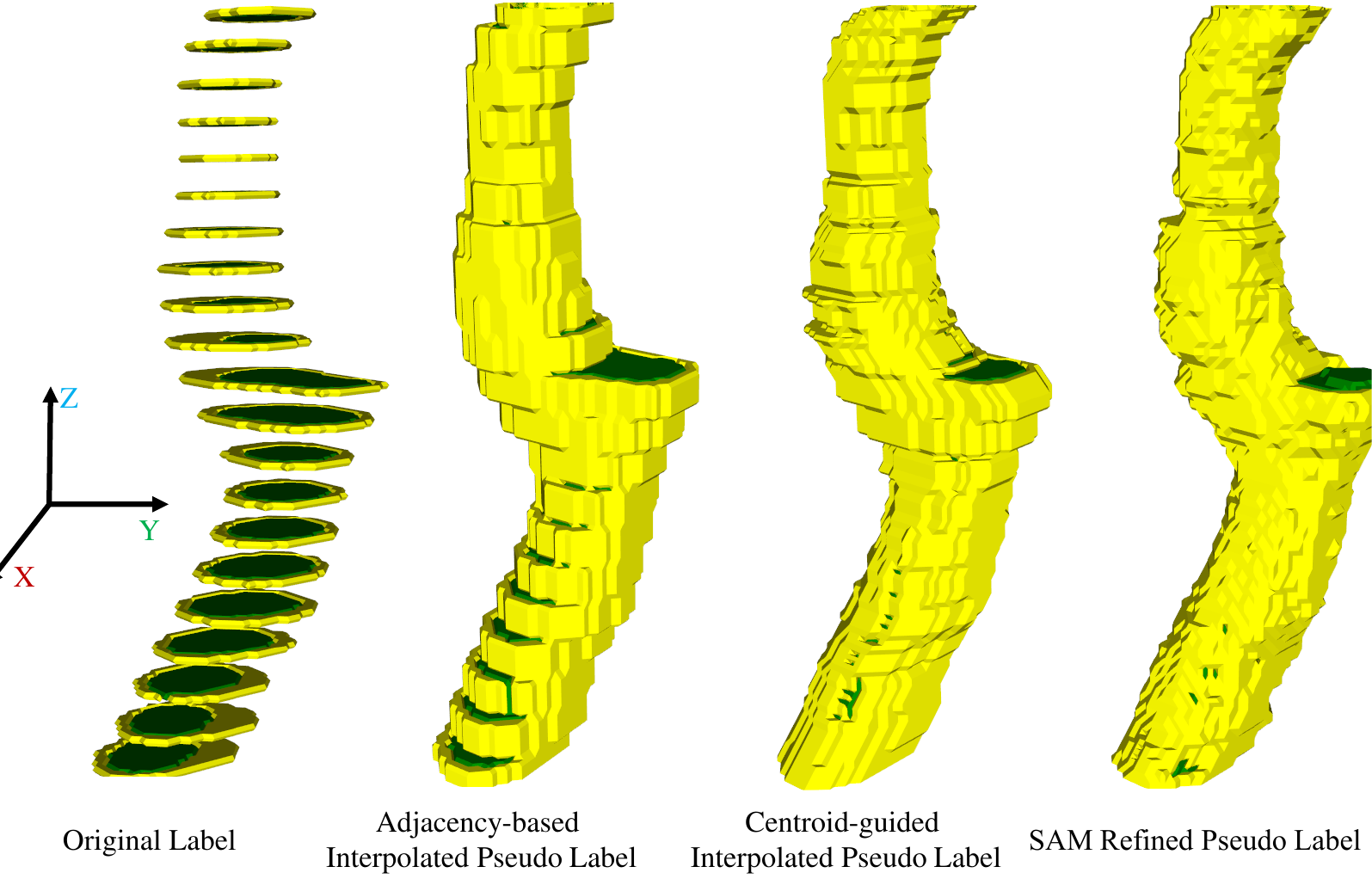}
\caption{Comparison of original label and three pseudo-label generation methods (A-IPL, C-IPL, and S-RPL).}
\label{fig:vessel3D}
\end{figure}
Taking the COSMOS2022~\cite{grand_challenge_vesselwall} dataset as an example, which provides only partial slice annotations in the spatial domain, we illustrate in Fig.~\ref{fig:vessel3D} a comparison between the original label and three pseudo-label generation approaches. A straightforward approach is training models solely on the available annotated slices. However, the lack of annotations in numerous slices leads to poor model recall, severely limiting its clinical value.

\begin{figure}[htbp]
\centering
\includegraphics[width=0.47\textwidth]{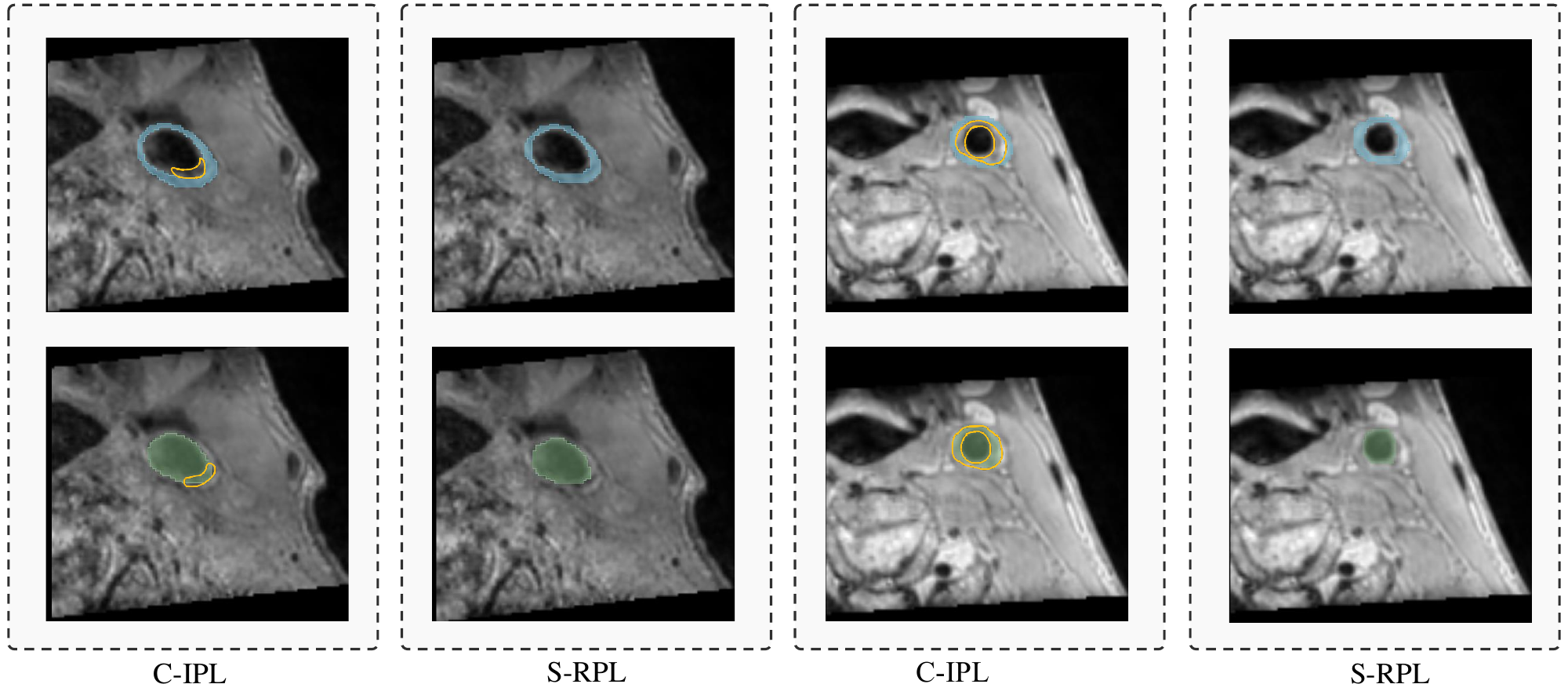}
\caption{Original label and three pseudo-label generation approaches}
\label{fig:vessel2D}
\vspace{-10pt}
\end{figure}
To enhance model performance, researchers initially proposed the Adjacency-based Interpolated Pseudo Label (A-IPL) method~\cite{hu2022label}, which directly copies labels from annotated slices to adjacent unlabeled ones. While this approach provides some spatial continuity, it generates problematic stair-like structures and exposed vessel lumens, contradicting the anatomical principle that vessel walls should consistently enclose the lumen. To address these limitations, we developed the Centroid-guided Interpolated Pseudo Label (C-IPL) method, which improves vessel annotation continuity by interpolating between centroid positions of adjacent annotated slices. 

Nevertheless, in regions with dramatic vessel morphology changes, the C-IPL method may generate suboptimal pseudo-labels as shown in Fig.~\ref{fig:vessel2D}, where vessel walls and lumens are represented in blue and green respectively. While C-IPL exhibits notable deviations due to rapid morphological changes (highlighted by yellow contours), our proposed SAM Refined Pseudo Label (S-RPL) method, which leverages SAM-Med2D's robust medical image generalization capabilities, accurately captures local vessel characteristics and generates anatomically plausible annotations. This approach not only maintains spatial continuity but also provides higher-quality supervision for segmentation model training.

The main contributions of this work are summarized as follows:

\begin{itemize}
\item We propose S-RPL, a SAM Refined Pseudo Label method that effectively handles vessel morphology variations at bifurcations and pathological regions. This method generates anatomically consistent annotations, serving as reliable supervision for network training.

\item We design DBF-UNet, an enhanced U-shaped architecture with three innovative modules: DSDBlock for efficient feature downsampling through dense connectivity, MLKBlock with statistical attention mechanisms for multi-scale feature learning, and BFFBlock for bidirectional feature fusion between different encoder levels, collectively enabling precise vessel delineation.

\item Extensive evaluations on the COSMOS2022 dataset validate our method's superiority over state-of-the-art approaches, demonstrating remarkable improvements in segmentation accuracy, particularly in challenging regions with complex vessel structures and limited annotations.
\end{itemize}

\section{RELATED WORKS}
\subsection{Segment Anything Model}
Large-scale vision foundation models represented by SAM (Segment Anything Model)~\cite{kirillov2023segment} have demonstrated remarkable zero-shot generalization capabilities and prompt-based segmentation performance in recent years. However, direct application of SAM to medical image segmentation exhibits notable limitations due to the significant domain gap between medical and natural images. To address this challenge, researchers have proposed various improved approaches: Med-SA~\cite{wang2023review} introduces a lightweight adapter module that achieves performance improvement by updating only 2\% of the parameters; SAM-Med2D~\cite{cheng2023sam} adopts a more comprehensive strategy by constructing a large-scale 2D medical image dataset and fine-tuning SAM to support multiple interactive prompting modes; while for 3D medical images, SAM-Med3D~\cite{wang2023sam} abandons SAM's pre-trained weights and employs a fully 3D network architecture, achieving superior segmentation performance through a two-stage training strategy on an extensive 3D dataset (22K images and 143K masks).
\vspace{-1.5pt}
\subsection{Carotid Artery Segmentation}
With the advancement of deep learning, carotid artery segmentation techniques have evolved significantly. Alblas et al. proposed a two-stage approach that first utilizes 3D UNet to predict arterial centerline distribution, followed by CNN-based wall thickness prediction in a polar coordinate system, ensuring the annular structure of vessel walls~\cite{alblas2022deep}. Subsequently, Hu et al. introduced a label propagation-based two-stage segmentation network, which initially obtains continuous 3D pseudo-labels through interpolation from limited 2D annotations, using these pseudo-labels to generate refined labels for subsequent nnUNet training~\cite{hu2022label}. More recently, Li et al. developed the MT-Net framework combining SAM and cross-modal transfer learning, which transfers vessel features learned from CTA data to MRI data while utilizing SAM to generate pseudo-labels for unlabeled slices, further enhancing segmentation performance~\cite{li2024carotid}.

Despite these efforts, existing methods face three critical limitations: (1)~Pseudo-labels from methods like A-IPL contain considerable noise when propagating between adjacent slices, particularly in complex regions like bifurcations; (2)~SAM-based approaches suffer from both the domain gap between natural and medical images and restricted prompt usage, as current methods only employ centroid-based point prompts while neglecting box and mask prompts; (3)~Ineffective multi-modal data utilization due to the scarcity of paired medical datasets.
\vspace{-3pt}
\section{Method}
\subsection{Stage 1 : Generating Pseudo Labels Based on SAM-Med2D}
To address these challenges, we propose a vessel segmentation method based on SAM-Med2D fine-tuning, aiming to generate high-quality pseudo-labels. The overall workflow is as follows :
\vspace{-5pt}
\begin{figure}[htbp]
\centering
\includegraphics[width=0.48\textwidth]{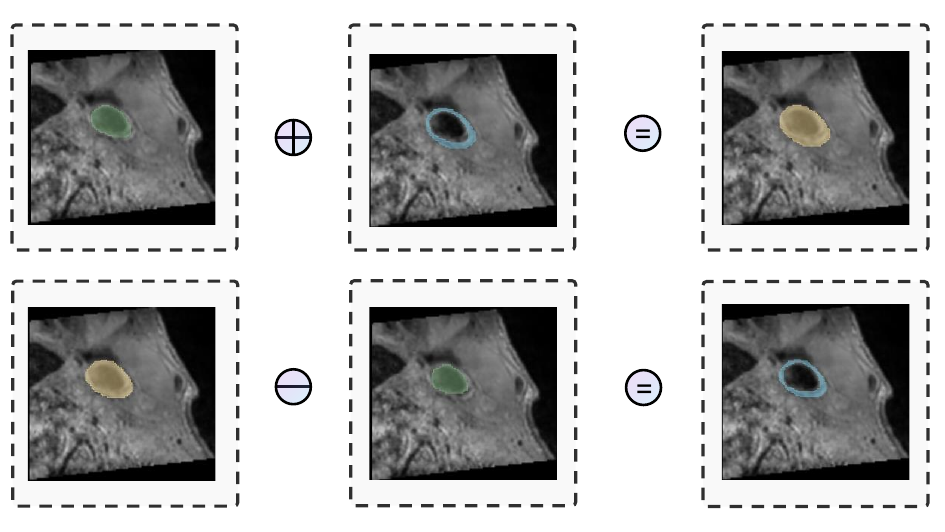}
\caption{Simplified vessel wall segmentation workflow. Vessel lumen and wall annotations are merged into complete vessel masks (top). The vessel wall mask is obtained by subtracting the lumen mask from the complete vessel mask (bottom).}
\label{fig:DataCons}
\end{figure}
\vspace{-15pt}
\begin{figure}[htbp]
\centering
\includegraphics[width=0.5\textwidth]{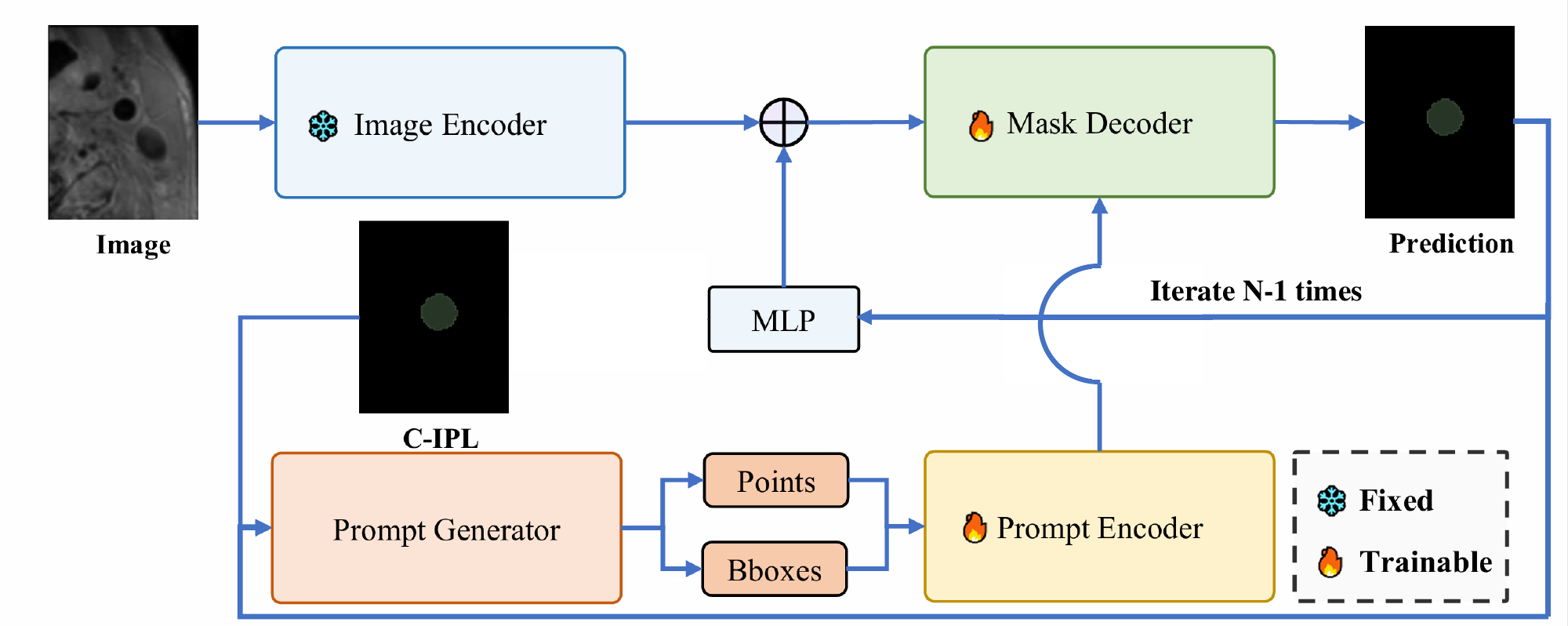}
\vspace{-10pt}
\caption{Fine-tuning framework of SAM-Med2D. The architecture uses three prompt types: Points, Bboxes and Masks. Initially, a random foreground point or bounding box from C-IPL serves as the sparse prompt. In later iterations, the framework identifies error regions between Prediction and C-IPL to generate new prompts. Previous predictions as mask prompts in subsequent iterations. The process refines segmentation results over N-1 iterations.}
\label{fig:SAMft}
\end{figure}
\vspace{-15pt}
\begin{figure}[htbp]
\centering
\includegraphics[width=0.47\textwidth]{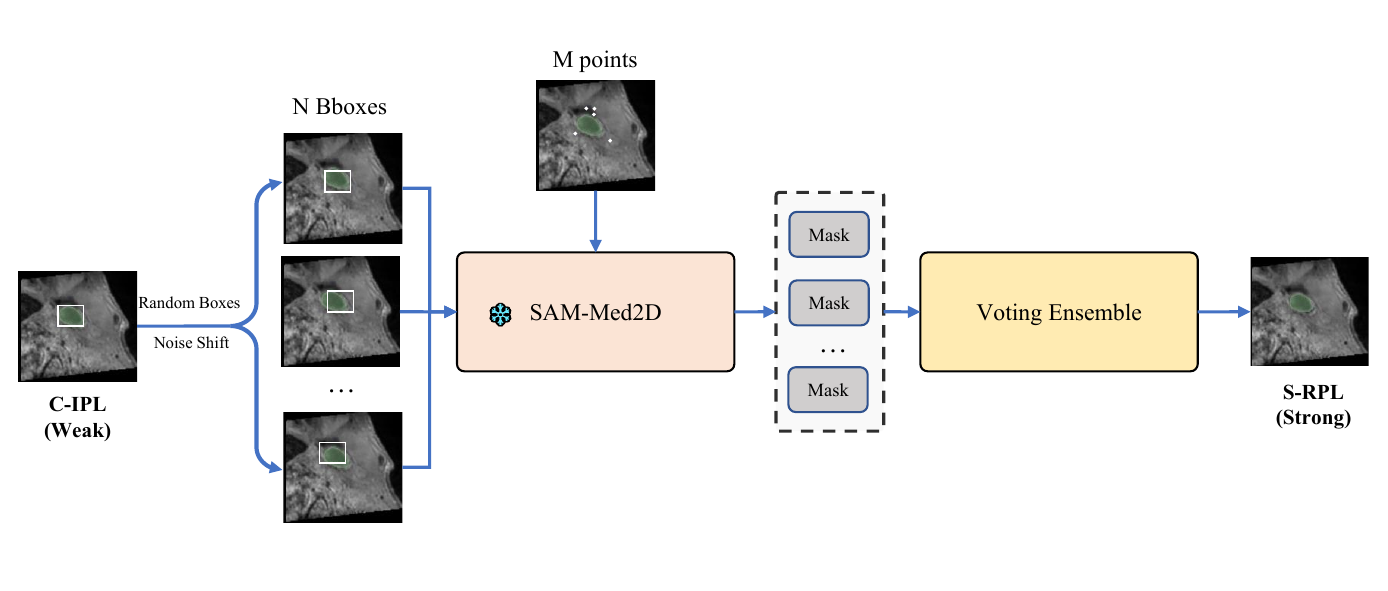}
\vspace{-10pt}
\caption{Inference optimization of SAM-Med2D}
\label{fig:SAMInfer}
\end{figure}
\begin{figure*}[htbp]
\centering
\includegraphics[width=1.0\textwidth]{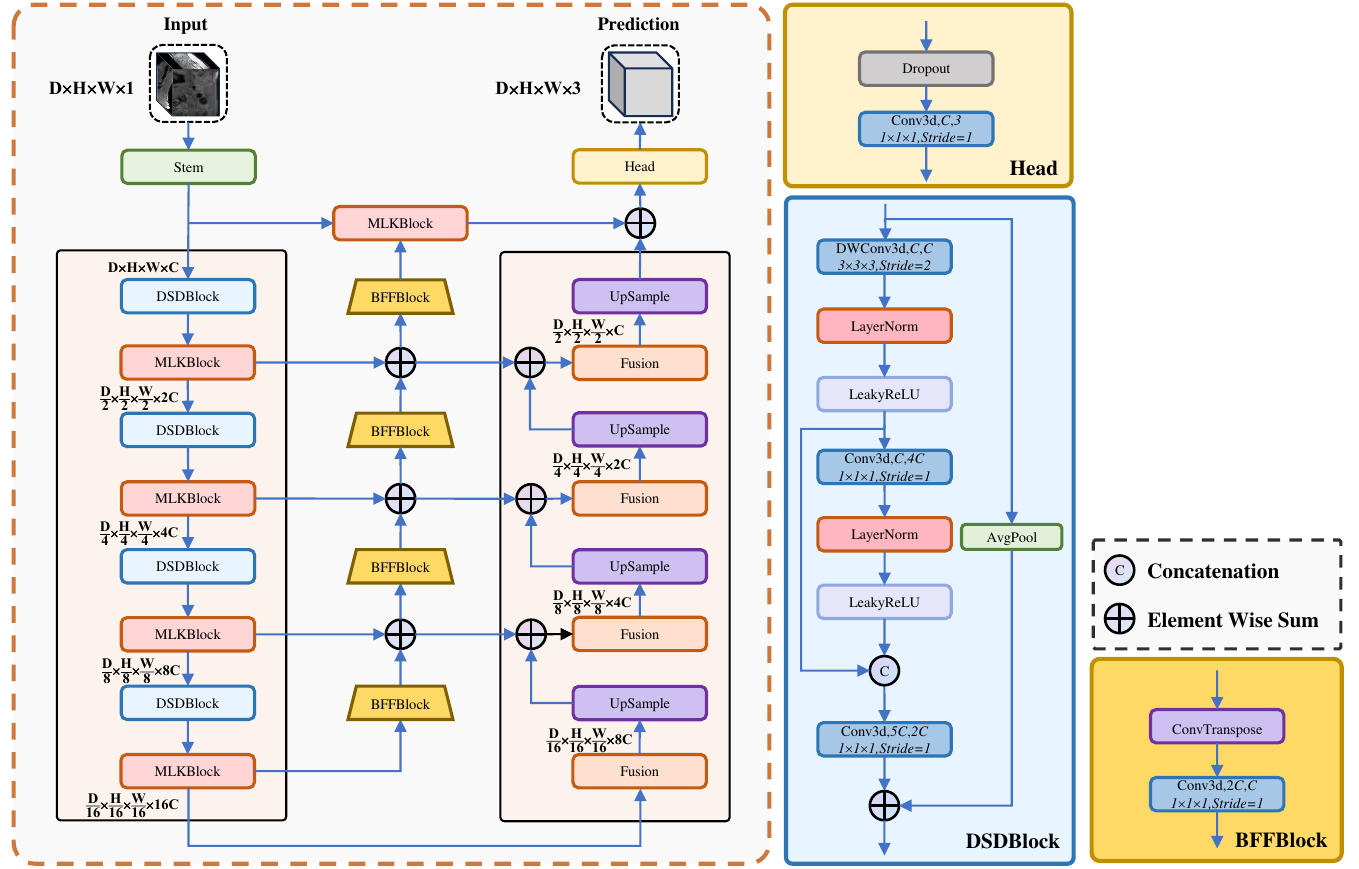}
\vspace{-3pt}
\caption{The architecture of DBF-UNet. The main structure (left) follows a U-shaped design where: (1) The encoder path employs DSDBlocks for downsampling and MLKBlocks for feature extraction; (2) BFFBlocks serve as enhanced skip connections, facilitating bidirectional information flow between encoder levels; (3) The decoder integrates deep features with the enhanced skip connections from the encoder. The right illustrates the detailed structures of Head, DSDBlock, and BFFBlock.}
\label{fig:NetworkArc}
\end{figure*}
\subsubsection{A Centroid-guided Interpolation Pseudo Label (C-IPL) generation method}
First, we computes vessel centroids from annotated slices in 3D labels and interpolates these centroid points to construct vessel centerlines. Subsequently, the labels from annotated slices are propagated along the centerlines to adjacent unlabeled slices, thereby generating spatially continuous 3D annotations.
\subsubsection{Dataset Construction and Label Processing for SAM-Med2D}
The expert-annotated slices are utilized for training, while slices with C-IPL generated labels serve as the test set. Given the complexity of vessel wall segmentation, we adopt a simplification strategy by merging vessel lumen and wall annotations into complete vessel masks during fine-tuning. The vessel wall mask is then obtained by subtracting the predicted lumen mask from the predicted complete vessel mask, as illustrated in Fig.~\ref{fig:DataCons}.
\begin{figure*}[htbp]
\centering
\includegraphics[width=1.0\textwidth]{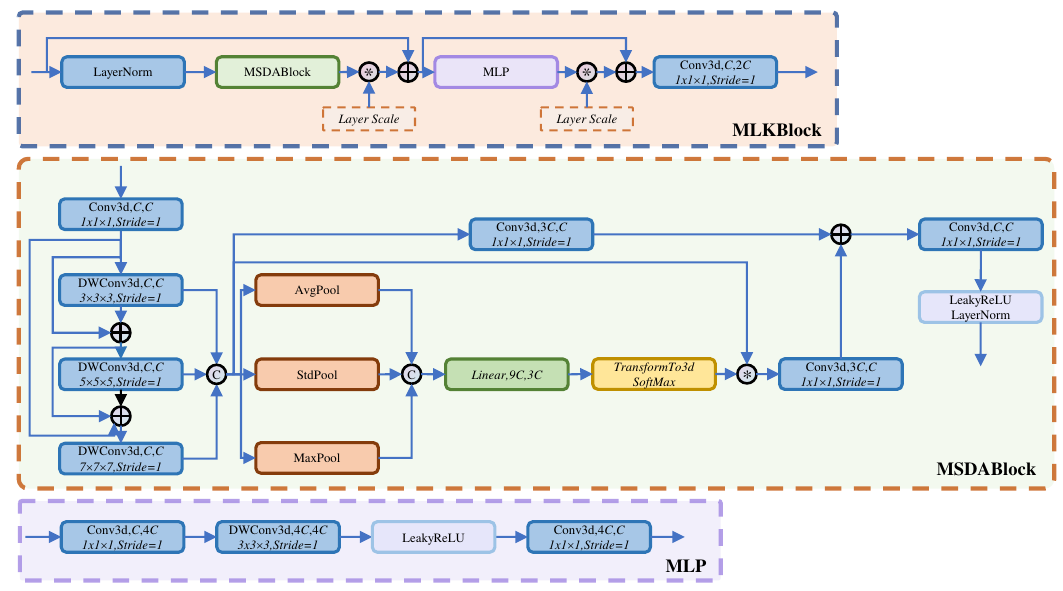}
\vspace{-3pt}
\caption{Detailed architecture of MLKBlock and its components. Top: The Transformer-inspired MLKBlock structure consisting of LayerNorm, MSDABlock, and feed-forward network. Middle: The MSDABlock architecture showing multi-scale feature extraction with different kernel sizes and statistical attention computation. Bottom: The MLP module design incorporating feed-forward layers and activation functions. Each component is equipped with layer scale parameters and residual connections for effective feature learning.}
\label{fig:MLKBlock}
\end{figure*}
\subsubsection{Model Fine-tuning}
To adapt SAM-Med2D for carotid artery segmentation, we separately fine-tune the model for vessel wall and lumen segmentation using expert-annotated images. We freeze the Image Encoder parameters and optimize only the Mask Decoder and Prompt Encoder through a multi-prompt strategy that combines point, box, and mask prompts. This approach enhances the model's capability to capture vessel details, as shown in Fig.~\ref{fig:SAMft}.
\subsubsection{Model Inference Optimization}
Due to the inherent limitations of C-IPL, vessel annotations may exhibit spatial deviations from their actual anatomical positions, particularly in regions with significant morphological variations. To address this challenge and enhance the robustness of our segmentation framework, we propose a comprehensive inference strategy combining noise-based box perturbation with an ensemble voting mechanism.

During inference, while C-IPL provides initial prompt information, we generate multiple perturbed versions of each interpolated bounding box to account for potential spatial deviations. These perturbed boxes produce diverse segmentation masks, which are then integrated through voting to generate SAM Refined Pseudo Labels (S-RPL). Subsequently, S-RPL and expert annotations are fused to create spatially continuous 3D labels for network training. The complete inference workflow is illustrated in Fig.~\ref{fig:SAMInfer}.

The key to this strategy lies in our adaptive noise injection scheme for box perturbation. Consider an original bounding box $B = \{(x_0, y_0, x_1, y_1)\}$. The perturbation process adapts to the target size through the following steps:

First, we calculate a size-dependent standard deviation:
\begin{equation}
    \sigma = \min(w, h) \times s
    \label{eq:std}
\end{equation}
where $w = |x_1 - x_0|$ and $h = |y_1 - y_0|$ represent the box dimensions, and $s$ is a predefined coefficient.

To ensure controlled perturbation, we bound the maximum noise amplitude:
\begin{equation}
    \delta = \min(M, 5\sigma)
    \label{eq:noise_bound}
\end{equation}
where $M$ constrains the maximum allowable perturbation.

The random perturbations are then sampled from:
\begin{equation}
    \varepsilon_x, \varepsilon_y \sim \mathcal{U}(-\delta, \delta)
    \label{eq:noise}
\end{equation}
where $\mathcal{U}(-\delta, \delta)$ denotes a uniform distribution over $[-\delta, \delta]$.

Finally, the perturbed box coordinates are computed as:
\begin{equation}
    B' = \{(x_0 + \varepsilon_x, y_0 + \varepsilon_y, x_1 + \varepsilon_x, y_1 + \varepsilon_y)\}
    \label{eq:perturbed_box}
\end{equation}
This adaptive perturbation mechanism ensures that the noise magnitude scales proportionally with the target size, while the ensemble voting of multiple predictions effectively compensates for potential C-IPL annotation deviations, enhancing the overall segmentation reliability.
\subsection{Stage 2 : DBF-UNet}
To enhance the accuracy and robustness of vessel segmentation, we propose the Dense Bidirectional Feature-fusion UNet (DBF-UNet). As illustrated in Fig.~\ref{fig:NetworkArc}, the core architecture of DBF-UNet consists of three innovative modules: Dense Spatial Downsampling Block (DSDBlock), Multi-level Kernel Block (MLKBlock), and Bidirectional Feature Fusion Block (BFFBlock).

\subsubsection{DSDBlock}
Inspired by the efficient dense connectivity design in RDNet~\cite{kim2025densenets}, we propose DSDBlock for feature downsampling, which enhances feature representation through progressive feature aggregation. For an input feature map $x\in\mathbb{R}^{B\times C\times D\times H\times W}$, DSDBlock applies a $3\times3\times3$ depthwise separable convolution with stride 2 for downsampling:
\begin{equation}
    x_1 = \text{DWConv}(x)
\end{equation}
Subsequently, a 1×1×1 pointwise convolution is employed to expand the channel dimension to quadruple the number of input channels.
\begin{equation}
    x_2 = \text{PWConv}(x_1)
\end{equation}
To effectively capture and integrate multi-scale feature information, we first concatenate all intermediate features along the channel dimension, followed by a pointwise convolution. A residual connection based on average pooling is introduced to maintain feature consistency:
\begin{equation}
    y = \text{PWConv}([x_1, x_2])+\text{AvgPool}(x)
\end{equation}
This dense connectivity structure combines depthwise separable convolution and pointwise convolution to achieve efficient feature reuse while maintaining computational efficiency.
\subsubsection{MLKBlock}
To enhance the multi-scale feature representation capability of the encoder, we design a Transformer-like Multi-level Kernel (MLK) module. As shown in Fig.~\ref{fig:MLKBlock}, the MLK module draws inspiration from Transformer architecture and consists of three core components: LayerNorm for feature normalization, Multi-scale Statistical Dense Attention (MSDA) block for capturing multi-scale features, and a feed-forward network comprising MLP and pointwise convolution.
The computation process of the MLK module can be formally expressed as:
\begin{equation}
\vspace{-2pt}
\begin{aligned}
x' &= \text{MSDA}(\text{LN}(x)) \\
z &= x + \gamma_1 \cdot x' \\
y &= \text{PWConv}(z + \gamma_2 \cdot \text{MLP}(\text{LN}(z)))
\end{aligned}
\end{equation}
where $\gamma_1$ and $\gamma_2$ are learnable Layer Scale parameters for adaptive feature fusion weight adjustment. Each component is equipped with Layer Scale parameters and residual connections, effectively preserving the advantages of Transformer architecture in feature extraction.
\subsubsection{MSDABlock}
The Multi-scale Statistical Dense Attention (MSDA) module serves as the core component of MLK Block. This module enhances feature representation through depthwise separable convolutions and statistical feature analysis, effectively leveraging both the complementarity of multi-scale features and their statistical properties. MSDA module comprises two key processing streams: densely connected multi-scale feature extraction and statistics-based attention computation.
In the multi-scale feature extraction phase, the module employs depthwise separable convolutions with different kernel sizes and progressively fuses features through dense connectivity. This progressive feature fusion strategy effectively captures spatial context information at different receptive fields, which can be expressed as:
\begin{equation}
\begin{aligned}
\vspace{-3pt}
    F_1 &= \text{Conv}_3(x) \\
    F_2 &= \text{Conv}_5(F_1) + F_1 \\
    F_3 &= \text{Conv}_7(F_2) + F_2 + F_1
\end{aligned}
\vspace{-1pt}
\end{equation}

where $\text{Conv}_k$ represents convolution layer with kernel size $k\times k\times k$. Through the design of residual connections, shallow-layer features can be directly propagated to deeper layers, effectively mitigating the gradient vanishing problem in deep network training.

In the statistical analysis phase, the module first concatenates multi-scale features along the channel dimension, then extracts statistical features and computes attention weights:
\begin{equation}
\begin{aligned}
\vspace{-6pt}
    F &= \text{Concat}({F_1, F_2, F_3}) \\
    S &= {\mu(F), \sigma(F), \text{max}(F)} \\
    A &= \text{Softmax}(W\cdot\text{Concat}(S))
\end{aligned}
\vspace{-1pt}
\end{equation}

Here, $F$ represents the concatenated multi-scale features, and $S$ comprises three crucial statistical measures: mean $\mu(F)$ captures distribution trends, standard deviation $\sigma(F)$ describes feature dispersion, and maximum value $\text{max}(F)$ preserves salient feature information. These statistical measures are transformed into attention weights $A$ through a learnable linear transformation matrix $W$ and softmax function.

The final feature fusion and output computation is represented as:
\vspace{-3pt}
\begin{equation}
    Y = \text{LN}(\text{ACT}(\text{Conv}(\text{Conv}(F\cdot A) + \text{Conv}(F))))
\end{equation}

where $F$ represents input features, $A$ denotes attention weights, LN and ACT indicate layer normalization and activation function respectively.

\subsubsection{BFFBlock}
To enhance feature extraction effectiveness, we propose the Bidirectional Feature Fusion (BFF) Block. This module achieves comprehensive feature representation by fusing bottom-up spatial details and top-down semantic information from the encoder. Specifically, the BFF module establishes bidirectional information flow between different encoder levels, facilitating effective interaction of multi-level features.

At layer $i$, the feature fusion process of the BFF module can be formulated as:
\begin{equation}
\begin{aligned}
\vspace{-3pt}
    F_i &= \text{BFF}(E_{out}^i) + E_{out}^{i-1} \\
    D_{in}^{i-1} &= F_i + D_{out}^{i}
\end{aligned}
\end{equation}
where $E_{out}^i$ denotes the output features from encoder layer $i$, $F_i$ represents the fused features processed by the BFF module, $D_{out}^{i}$ and $D_{in}^{i-1}$ indicate the output features from decoder layer $i$ and input features to decoder layer $i-1$ respectively. Through the design of residual connections, the module better preserves original feature information while integrating multi-level feature representations. In shallow encoder stage, the original block is replaced by MLKBlock, which process the combined features from both the shallowest encoder output and the BFFBlock through element-wise addition.
\section{EXPERIMENTS}
\subsection{Dataset}
Experimental validation is conducted on the COSMOS2022~\cite{grand_challenge_vesselwall} dataset, which comprises carotid MRI scans from 50 subjects. The images are acquired using a Philips 3T MRI system with a 3D VISTA sequence, achieving isotropic voxels with 0.6mm resolution in all dimensions. The expert annotations in this dataset follow an interval sampling strategy, where vessel wall and lumen boundaries are annotated at regular intervals along the axial direction of the 3D volume.
\subsection{Implementation Details}
In our experimental design, the COSMOS dataset is split into training, validation, and testing sets with a ratio of 35:2:13. For single stage methods, A-IPL are used for training. In our two stage framework, we employ Centroid-guided Interpolated Pseudo Labels for SAM-Med2D fine-tuning. For SAM-Med2D fine-tuning, we configure the following parameters: input image resolution of $256\times256$, batch size of 16, Adam optimizer with an initial learning rate of $1\times10^{-4}$, and 200 training epochs. The loss function combines Focal Loss, Dice Loss, and IoU Loss with relative weights of 20:1:1. During inference, 10 candidate boxes are generated for each initial bounding box through noise perturbation.

To ensure fair comparison, all methods are trained with consistent configurations: 3D patch size of $128\times128\times128$, batch size of 2, Adam optimizer with an initial learning rate of $5\times10^{-4}$, and 300 epochs. The loss function utilizes a weighted combination of cross-entropy loss and Dice Loss.
\begin{table}[!h]
\centering
\renewcommand{\arraystretch}{1.2}
\caption{Performance Comparison of Different Methods on Vessel Lumen Segmentation on COSMOS2022 Dataset}
\setlength{\tabcolsep}{6.3pt}
\begin{tabular}{c|ccccc}
\hline
\textbf{Method} & \textbf{Dice} & \textbf{IoU} & \textbf{Pre} & \textbf{Rec} & \textbf{ASD} \\
\hline
nnUNet~\cite{isensee2021nnu} & 92.52 & 88.23 & 94.71 & 91.08 & 0.1755 \\
UNETR~\cite{hatamizadeh2022unetr} & 87.13 & 81.83 & 90.99 & 86.01 & 0.5350 \\
SwinUNETR~\cite{hatamizadeh2021swin} & 90.86 & 86.24 & 91.17 & 92.09 & 0.3177 \\
MedNeXt~\cite{roy2023mednext} & 88.70 & 83.59 & 92.67 & 87.42 & 0.4488 \\
UX-Net~\cite{lee20223d} & 90.64 & 85.51 & 92.91 & 89.92 & 0.3247 \\
LKM-UNet~\cite{wang2024lkm} & 93.06 & 88.90 & 94.77 & 92.46 & 0.1906 \\
UMamba~\cite{ma2024u} & 92.76 & 88.34 & 94.85 & 91.65 & 0.1825 \\
Label Propogation~\cite{hu2022label} & 93.85 & 89.74 & 94.53 & 93.63 & 0.3066 \\
DBF-UNet & 94.61 & 90.43 & 94.72 & 95.13 & 0.2072 \\
SAM-Med2D+DBF-UNet & \textbf{95.22} & \textbf{91.03} & \textbf{95.78} & \textbf{95.19} & \textbf{0.1618} \\
\hline
\end{tabular}
\label{tab:CompExpLumen}
\end{table}
\vspace{-10pt}
\begin{table}[!h]
\centering
\renewcommand{\arraystretch}{1.2}
\setlength{\tabcolsep}{6.3pt}
\caption{Performance Comparison of Different Methods on Vessel Wall Segmentation on COSMOS2022 Dataset}
\setlength{\tabcolsep}{5pt}
\begin{tabular}{c|ccccc}
\hline
\textbf{Method} & \textbf{Dice} & \textbf{IoU} & \textbf{Pre} & \textbf{Rec} & \textbf{ASD} \\
\hline
nnUNet & 85.27 & 75.26 & 85.28 & 86.85 & 0.2328  \\
UNETR & 79.04 & 68.05 & 81.52 & 79.36 & 0.5908\\
SwinUNETR & 82.76 & 72.69 & 84.85 & 83.55 & 0.3423  \\
MedNeXt & 80.04 & 69.18 & 82.59 & 80.27 & 0.5187  \\
UX-Net & 81.65 & 71.65 & 84.48 & 81.73 & 0.3439 \\
LKM-UNet & 84.85 & 75.33 & 85.38 & 86.01 & 0.2269 \\
UMamba & 85.10 & 74.76 & 86.03 & 85.42 & 0.2205 \\
Label Propogation & 85.56 & 76.18 & 85.44 & \textbf{87.43} & 0.2767 \\
DBF-UNet & 85.93 & 75.98 & 86.52 & 86.61 & 0.2378 \\
SAM-Med2D+DBF-UNet & \textbf{86.08} & \textbf{76.48} & \textbf{86.55} & 87.09 & \textbf{0.1937}\\

\hline
\end{tabular}
\vspace{-8pt}
\label{tab:CompExpWall}
\end{table}
\subsection{Evaluation Metrics}
To comprehensively evaluate the segmentation performance, we employ five widely-used metrics:

\textbf{Dice Similarity Coefficient (DSC)} measures the overlap between the predicted segmentation and ground truth:
\begin{equation}
    \text{DSC} = \frac{2|X \cap Y|}{|X| + |Y|} = \frac{2\text{TP}}{2\text{TP} + \text{FP} + \text{FN}}
\end{equation}
where $X$ and $Y$ represent the predicted and ground truth segmentation masks respectively.

\textbf{Intersection over Union (IoU)} evaluates the ratio between the intersection and union of prediction and ground truth:
\begin{equation}
    \text{IoU} = \frac{|X \cap Y|}{|X \cup Y|} = \frac{\text{TP}}{\text{TP} + \text{FP} + \text{FN}}
\end{equation}

\textbf{Precision} quantifies the proportion of correctly predicted positive pixels among all predicted positive pixels:
\begin{equation}
    \text{Precision} = \frac{\text{TP}}{\text{TP} + \text{FP}}
\end{equation}

\textbf{Recall} measures the proportion of correctly identified positive pixels among all actual positive pixels:
\begin{equation}
    \text{Recall} = \frac{\text{TP}}{\text{TP} + \text{FN}}
\end{equation}

\textbf{Average Surface Distance (ASD)} calculates the mean distance between the boundaries of prediction and ground truth:
\begin{equation}
    \text{ASD}(X,Y) = \frac{\sum_{x \in B_X}d_{\min}(x,B_Y) + \sum_{y \in B_Y}d_{\min}(y,B_X)}{|B_X| + |B_Y|}
\end{equation}
where $B_X$ and $B_Y$ denote the boundary point sets of prediction and ground truth respectively, and $d_{\min}(p,S)$ represents the minimum Euclidean distance from point $p$ to point set $S$.

These metrics complement each other in evaluating different aspects of segmentation quality: DSC and IoU assess overall segmentation accuracy, Precision and Recall provide insights into false positive and false negative predictions, while ASD specifically evaluates boundary accuracy.
\subsection{Comparative Experiment}

To comprehensively evaluate the effectiveness of our proposed method, we conduct extensive experiments on the COSMOS2022 dataset for both vessel lumen and wall segmentation tasks. We compare our approach with several state-of-the-art medical image segmentation methods, including convolution-based architectures (nnUNet, UX-Net, MedNeXt), transformer-based approaches (UNETR, SwinUNETR), Mamba-based methods (LKM-UNet, UMamba), and a two-stage segmentation framework that utilizes nnUNet for Label-Propagation. In Tables \ref{tab:CompExpLumen} and \ref{tab:CompExpWall}, we present the quantitative comparison results, where bold numbers indicate the best performance among all compared methods.

As shown in Table \ref{tab:CompExpLumen}, our proposed SAM-Med2D+DBF-UNet framework achieves superior performance in vessel lumen segmentation across all evaluation metrics. Specifically, it attains the highest Dice score of 95.22\% and IoU of 91.03\%, surpassing the baseline nnUNet by 2.70\% and 2.80\% respectively. The precision (95.78\%) and recall (95.19\%) metrics further demonstrate our method's capability in reducing both false positives and false negatives. Notably, our method achieves the best ASD value (0.1618), indicating more accurate boundary delineation.

For the more challenging vessel wall segmentation task (Table \ref{tab:CompExpWall}), our method maintains competitive performance with a Dice score of 86.08\% and IoU of 76.48\% and ASD of 0.1937. While the improvement margin is relatively smaller compared to the lumen segmentation task, our approach still outperforms other methods in most metrics.

Compared to transformer-based methods like SwinUNETR, our approach demonstrates substantial improvements in both tasks. For instance, in lumen segmentation, we surpass SwinUNETR by 4.36\% in Dice score. The performance gap is even more pronounced in wall segmentation, where our method outperforms SwinUNETR by 3.32\%.
\begin{table}[!h]
\centering
\renewcommand{\arraystretch}{1.2}
\caption{Computational Complexity Comparison of Different Methods}
\setlength{\tabcolsep}{12pt}
\begin{tabular}{c|cc}
\hline
\textbf{Method} & \textbf{Parameters (M)} & \textbf{FLOPs (G)} \\
\hline

MambaClinix & 108.33 & 13960.54 \\
UNETR & 121.35 & 391.20 \\
LKM-UNet& 102.19 & 4938.70 \\
UMamba & 69.37 & 13447.52 \\
UX-Net & 34.71 & 1349.14 \\
nnUNet & 30.45 & 2610.66 \\
MedNeXt & 17.60 & 525.70 \\
SwinUNETR & 15.64 & 394.96 \\
DBF-UNet & \textbf{2.81} & \textbf{211.52} \\
\hline

\end{tabular}
\vspace{-5pt}
\label{tab:complexity}
\end{table}
\begin{table}[!h]
\centering
\renewcommand{\arraystretch}{1.2}
\setlength{\tabcolsep}{12pt}
\caption{Ablation Study of Key Components in DBF-UNet for Vessel Lumen Segmentation on COSMOS2022}
\setlength{\tabcolsep}{5pt}
\begin{tabular}{cc|ccccc}
\hline
\textbf{BFF} & \textbf{MSDABlock} & \textbf{Dice} & \textbf{IoU} & \textbf{Pre} & \textbf{Rec} & \textbf{ASD}\\
\hline
- & - & 93.41 & 88.91 & 94.64 & 93.01 & 0.2592 \\
\checkmark & - & 93.80 & 89.63 & 94.45 & 93.92 & 0.2042  \\
- & \checkmark & 94.10 & 89.93 & 94.67 & 94.27 & \textbf{0.1900}  \\
\checkmark & \checkmark & \textbf{94.61} & \textbf{90.43} & \textbf{94.72} & \textbf{95.13} & 0.2072\\
\hline
\end{tabular}
\vspace{-7pt}
\label{tab:AbaExpLumen}
\end{table}
\begin{table}[!h]
\centering
\renewcommand{\arraystretch}{1.2}
\caption{Ablation Study of Key Components in DBF-UNet for Vessel Wall Segmentation on COSMOS2022}
\setlength{\tabcolsep}{5pt}
\begin{tabular}{cc|ccccc}
\hline
\textbf{BFF} & \textbf{MSDABlock} & \textbf{Dice} & \textbf{IoU} & \textbf{Pre} & \textbf{Rec} & \textbf{ASD}\\
\hline
- & - & 84.57 & 74.33 & 85.43 & 85.29 & 0.2648 \\
\checkmark & - & 85.42 & 75.69 & \textbf{87.95} & 84.47 & 0.2512\\
- & \checkmark & 85.65 & 75.35 & 87.16 & 85.52 & 0.2449  \\
\checkmark & \checkmark &  \textbf{85.93} & \textbf{75.98} & 86.52 & \textbf{86.61} & \textbf{0.2378} \\
\hline
\end{tabular}
\label{tab:AbaExpWall}
\vspace{-5pt}
\end{table}

Table~\ref{tab:complexity} shows the computational efficiency comparison of DBF-UNet against existing methods. Our model contains only 2.81M parameters, which is significantly smaller than SwinUNETR (15.64M), UNETR (121.35M), and MambaClinix (108.33M). The computational cost of DBF-UNet (211.52 GFLOPs) is also notably lower than Mamba-based models (UMamba: 13447.52 GFLOPs, MambaClinix: 13960.54 GFLOPs), indicating its high computational efficiency while maintaining competitive performance.
\begin{figure}[htbp]
\centering
\includegraphics[width=0.47\textwidth]{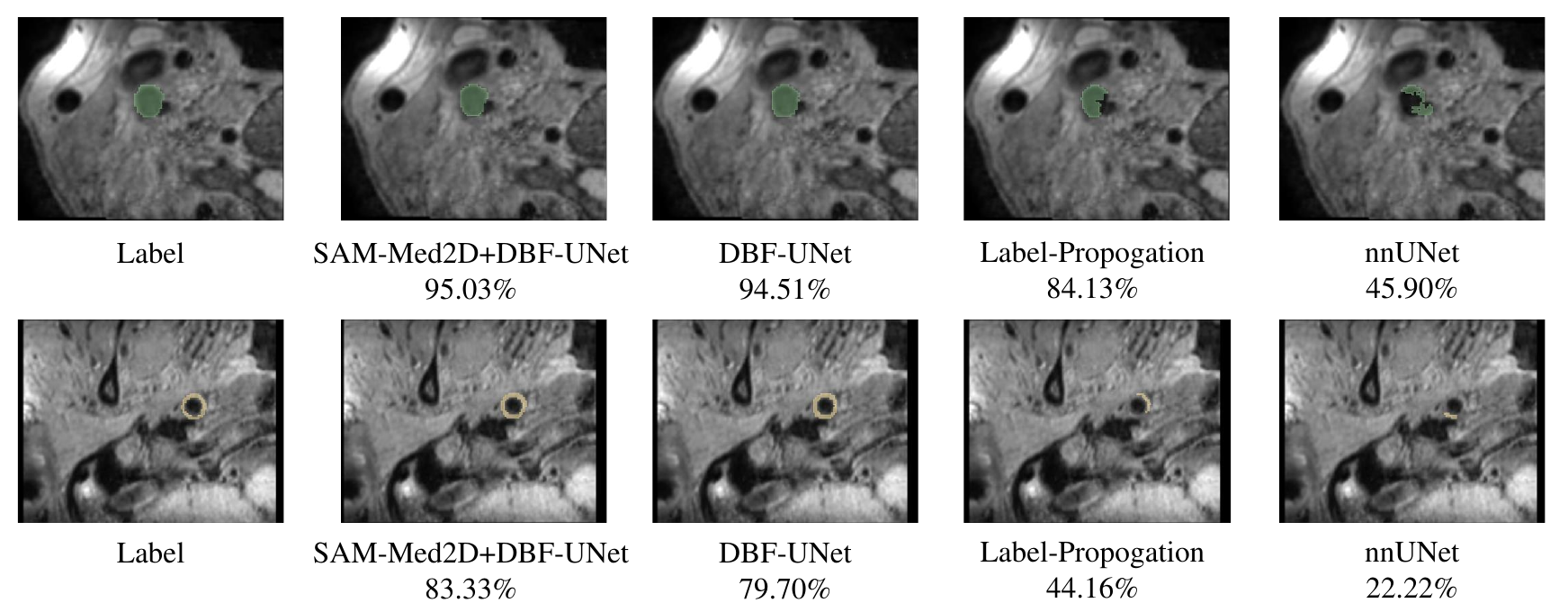}
\vspace{-2pt}
\caption{Visual comparison of segmentation results from four different methods for vessel lumen (top row) and vessel wall (bottom row), with corresponding Dice coefficients shown below each result.}
\label{fig:show}
\end{figure}

As illustrated in Fig.~\ref{fig:show}, we provide qualitative visualization results for vessel lumen (top row) and vessel wall (bottom row) segmentation. Our two-stage approach, SAM-Med2D+DBF-UNet, outperforms all other methods in both tasks, achieving Dice scores of 95.03\% and 83.33\%, respectively. These results represent significant improvements over the single-stage DBF-UNet, which scores 94.51\% and 79.70\%, particularly in the delineation of vessel walls. In the vessel wall segmentation task, both Label-Propagation and nnUNet suffer from severe under-segmentation, resulting in discontinuous and fragmented vessel walls across adjacent slices. This structural discontinuity not only undermines anatomical integrity but also adversely affects the reliability of downstream clinical analyses. In contrast, our proposed methods consistently preserve the structural continuity and precise boundary delineation of both vessel lumen and walls. Notably, SAM-Med2D+DBF-UNet excels in capturing fine-grained details of vessel walls.

\subsection{Ablation Experiment}
To validate the effectiveness of our proposed components in DBF-UNet, we conduct ablation experiments on vessel segmentation tasks, as shown in Tables \ref{tab:AbaExpLumen} and \ref{tab:AbaExpWall}. For lumen segmentation, incorporating BFFBlock improves the Dice score from 93.41\% to 93.80\%, while MSDABlock independently enhances it to 94.10\%. The integration of both components further boosts the performance to 94.61\%, demonstrating substantial improvements in both segmentation accuracy and boundary recall (Rec increases from 93.01\% to 95.13\%). In wall segmentation, BFFBlock and MSDABlock individually improve the Dice score from 84.57\% to 85.42\% and 85.65\%, respectively. When combined, these components achieve optimal performance with a Dice score of 85.93\% and enhance the recall rate from 85.29\% to 86.61\%. 

These results demonstrate that BFF enhances feature fusion capabilities, while MSDABlock strengthens spatial attention mechanisms. Their combination leads to complementary benefits, particularly evident in the improved recall rates for both segmentation tasks. The consistent performance gains across different metrics validate the necessity of both components in our architecture.
\section{CONCLUSION}
In this paper, we have presented a two-stage framework for carotid artery segmentation that effectively addresses the challenges of spatially discontinuous annotations and complex vessel morphology. Our method combines SAM-Med2D fine-tuning with centroid-guided interpolation, producing high-quality pseudo-labels through noise-based box perturbation and voting mechanisms. The proposed DBF-UNet, featuring DSDBlock for dense downsampling, MLKBlock for feature enhancement, and BFFBlock for bidirectional fusion, demonstrates superior capability in capturing vessel characteristics. Evaluated on the COSMOS2022 dataset, our method achieves state-of-the-art performance with Dice scores of 95.22\% and 86.08\% for vessel lumen and wall segmentation, respectively.

\bibliographystyle{IEEEbib}
\bibliography{ref}

\end{document}